\documentclass[sigconf,natbib=false,true,nonacm]{acmart}

\usepackage{tcolorbox}
\usepackage{todonotes}
\usepackage{tikz}
\usepackage{tabularx}
\usepackage[inline]{enumitem}
\usetikzlibrary{arrows.meta, positioning}
\usepackage{graphicx}
\usepackage{multirow}
\usepackage{array}
\usepackage{adjustbox}
\usepackage{amsmath}
\usepackage{amssymb}
\AtBeginDocument{%
  }

\copyrightyear{2026}
\acmYear{2026}
\acmDOI{XXXXXXX.XXXXXXX}
\acmConference[SAC'26]{The 41st ACM/SIGAPP Symposium on Applied Computing}{March 23--27, 2026}{Thessaloniki, Greece}
\acmISBN{979-X-XXXX-XXXX-X/26/03}

\RequirePackage[
  datamodel=acmdatamodel,
  style=acmnumeric,
  ]{biblatex}

\newcommand{\secref}[1]{\mbox{Section~\ref{#1}}}
\newcommand{\tabref}[1]{\mbox{Table~\ref{#1}}}
\newcommand{\figref}[1]{\mbox{Figure~\ref{#1}}}

\begin{document}

\title{Integrating Causal Reasoning into Automated Fact-Checking}
\subtitle{Extended Version}

\author{Youssra Rebboud}
\email{youssra.rebboud@eurecom.fr}
\orcid{0000-0003-3507-5646}
\affiliation{%
  \institution{EURECOM}
  \city{Sophia Antipolis}
  \country{France}
}

\author{Pasquale Lisena}
\email{pasquale.lisena@eurecom.fr}
\orcid{0000-0003-3094-5585}
\affiliation{%
  \institution{EURECOM}
  \city{Sophia Antipolis}
  \country{France}
}

\author{Raphael Troncy}
\email{raphael.troncy@eurecom.fr}
\orcid{0000-0003-0457-1436}
\affiliation{%
  \institution{EURECOM}
  \city{Sophia Antipolis}
  \country{France}
}


\begin{abstract}
In fact-checking applications, a common reason to reject a claim is to detect the presence of erroneous cause-effect relationships between the events at play. However, current automated fact-checking methods lack dedicated causal-based reasoning, potentially missing a valuable opportunity for semantically rich explainability. To address this gap, we propose a methodology that combines event relation extraction, semantic similarity computation, and rule-based reasoning to detect logical inconsistencies between chains of events mentioned in a claim and in an evidence. Evaluated on two fact-checking datasets, this method establishes the first baseline for integrating fine-grained causal event relationships into fact-checking and enhance explainability of verdict prediction.
\end{abstract}

\ccsdesc[500]{Information systems~Decision support systems}
\ccsdesc[500]{Computing methodologies~Information extraction}
\ccsdesc[500]{Computing methodologies~Causal reasoning and diagnostics}

\keywords{Fact-checking, Causal Reasoning, Explainability}

\begin{teaserfigure}
    \centering
    \adjustbox{margin=1em,width=\textwidth,frame,center,bgcolor=lightgray}{This paper is an extended version of the homonym paper accepted at ACM SAC (KNLP Track) and published in ACM proceedings.}
\end{teaserfigure}


\maketitle

\section{Introduction}
\label{sec:introduction}
The rapid proliferation of both accurate and misleading content on the Web has made automated fact-checking an essential task. While fact-checking spans a wide range of subtasks, a fundamental component is the assessment of entailment between a claim and its corresponding evidence. Existing models for textual entailment, often based on deep learning, are widely used for verdict prediction~\cite{guo-etal-2022-survey}. However, these models typically operate as black boxes, offering limited insight into their reasoning processes. To address this, prior efforts have explored explainability through attention mechanisms, summarization, or symbolic rule extraction~\cite{kotonya-toni-2020-explainable}.

Causal reasoning has recently gained traction as a means to support factual consistency within claims~\cite{chen2025counterfactual,si-etal-2024-checkwhy,tan-etal-2024-enhancing-fact}. These approaches rely on detecting cause-effect relationships between events described in the claim and those found in the evidence. However, entailment between claim and evidence is not solely based on a general notion of causality. In many real-world cases, the causal relationships between events are more nuanced
and contradictions arise not from the absence of a causal chain, but from deeper incompatibilities between relational semantics. Let us illustrate with the following example:\\
\textbf{Claim:} \textit{Taking the vaccine prevented infection}. \\
\textbf{Evidence:} \textit{The vaccine triggered an immune response that blocked the virus from spreading in the body}.

\noindent
Existing causal reasoning models typically represent relations only through general-purpose \textit{cause} links, without distinguishing fine-grained relations such as \textit{prevent}, \textit{enable}, or \textit{intend}. As such, they look for a direct match:
\[
\text{Vaccine} \xrightarrow{\text{cause}} \text{No infection}
\]
In this example, no such direct causal chain exists in the claim, which makes reasoning with a single causality relationship difficult. Moreover, many systems oversimplify causal relations by skipping intermediate steps. This leap from \(A\) (vaccine) to \(D\) (no infection) obscures the underlying logic and makes the reasoning path unclear to users. Considering the following notation:

\vspace{-10pt}
\[
\begin{array}{rlrl}
A: & \textit{Taking the vaccine} & 
C: & \textit{Immune response} \\
B: & \textit{Infection} & 
D: & \textit{Virus blocked}
\end{array}
\]

\noindent
and the following relations:
\begin{itemize}
    \item $A \xrightarrow{\text{causes}} C$ \hspace{0.7cm} (\textit{Vaccine causes immune response})
    \item $C \xrightarrow{\text{prevents}} B$ \hspace{0.5cm} (\textit{Immune response prevents infection})
\end{itemize}

\noindent
an automated systems should infer:
\[
A \xrightarrow{\text{causes}} C \xrightarrow{\text{prevents}} B \Rightarrow A \xrightarrow{\text{prevents}} B
\]

In this work, we propose a novel causal explanation-based verdict prediction system that relies on semantically-precise event relations -- namely \textit{cause}, \textit{prevent}, \textit{intend} and \textit{enable} -- derived from the FARO ontology~\cite{FARO}, and for which a dataset has been provided in~\cite{GPTaugment}. This system generates human-readable explanations based on causality extraction and a set of rules that identify the alignment or misalignment of claims and evidence. It is important to say that our approach only applies to claims that include at least one causal relation between events.

By integrating causal reasoning into the verdict prediction process, we address the limitations of existing explainability methods. The use of semantically defined relationships ensures that the explanations align with human reasoning. For this reason, our approach can also be used in combination with other fact-checking systems, with the final benefit of increasing the trust in the systems themselves. In summary, this work makes the following contributions:
\begin{enumerate}
    \item we propose a set of high-level reasoning rules for verdict prediction to be applied to causal relations;
    \item we develop a complete pipeline for applying those rules to sentences in claim-evidence pairs.
\end{enumerate}
All data, software and experiments are publicly available at \url{https://github.com/ANR-kFLOW/Fact_checking_reasoner}.


\section{Related Work}
\label{sec:related_work}
Numerous end-to-end fact-checking systems have been developed. In \cite{hassan}, the authors created a system that assesses claim veracity using keyword searches for evidence and knowledge bases, relying on traditional features such as part-of-speech tags and sentiment for a 3-way classification experimenting with classifiers such as random forests and SVMs. In contrast, others employ deep neural networks for evidence selection and natural language inference, marking early examples of explainable fact-checking~\cite{defend}. Popat et al.~\cite{popat-etal-2018-declare} use attention mechanism as a way to extract the most important words from an evidence as an explanation. The explanation gathered from the aforementioned ways is often not comprehensive \cite{kotonya-toni-2020-explainable}. Finally, other methods rely on summarization~\cite{atanasova-etal-2020-generating-fact}. 

Recently, some attention has been directed towards causal reasoning in fact checking~\cite{chen2025counterfactual,si-etal-2024-checkwhy}. In \cite{tan-etal-2024-enhancing-fact}, the use of LLMs for causal deductive reasoning is showcased, by leveraging causal graphs and counterfactual reasoning for fact verification. These approaches are limited to what we refer to as \textit{direct causality} within the evidence chain, without considering other types of event relations as in~\cite{FARO}. 

In~\cite{causal_intervetion}, a method for a multi-modal detection of fake news leverages causal intervention to identify and reduce psycholinguistic bias in textual content, while counterfactual reasoning is employed to isolate and address the bias introduced by image-only cues, simulating scenarios where only visual content is available.

To the best of our knowledge, no prior research has leveraged fine-grained event-level causal relations, which are semantically richer and more nuanced than the broad or vague notions of causality typically used in explainable systems.

\section{Reasoning Rules}
\label{sec:reasoning_rules}
In this section, we introduce relation-based reasoning rules to deduce the most probable verdict for a given claim, knowing the evidence. These rules are intended to be found: 
\begin{enumerate*}
    \item between events in the claim;
    \item between events in each evidence;
    \item between one event in the claim and one event in the evidence (or vice-versa).
\end{enumerate*}

\subsection{Types of Relationships}
The FARO ontology~\cite{FARO} defines a set of semantically precise event relations. This ontology not only provides a textual definition to several event relations, but already defines logical axioms such as transitivity or disjunction using the OWL representation language.

We focus our work on the four FARO relations, covering a broad range of cases: 
\begin{itemize}
    \item \textbf{direct-cause}: a relationship between an event and its effect. \\ 
    Ex: \textit{the earthquake has left behind dozens of deaths in Japan.}
    (earthquake, causes, deaths) \\
    \textbf{Transitive}\footnote{We adopt the commonly held assumption that direct causality is transitive. Despite ongoing debates~\cite{McDonnell2018}, we align with the prevailing perspective in the literature that supports this assumption.} |  \textbf{Disjointwith}: does-not-cause.

    \item \textbf{prevents}: connect an event instance with the event for which is the cause of not happening.\\
    Ex: \textit{Boy Scouts of America are committed to act against racial injustice.} 
    (act, prevents, racial injustice) \\
    \textbf{SuperProperty}: does-not-cause |  
    \textbf{Disjointwith}: cause
    
    \item \textbf{intends-to-cause}: establish a link between an event and its intended effect, regardless of whether the desired outcome is ultimately achieved. \\
    Ex: \textit{The company implemented a marketing strategy to boost the product sales.}
    (implementing, intends to cause, boost)\\
    \textbf{Disjointwith}: does-not-cause, prevent.

    \item \textbf{enables}: connect a condition with the event it is contributing to perform as an enabling factor.\\
    Ex: \textit{having access to reliable internet grants students access to online courses.} 
    (access to reliable internet, enables, access to online courses) \\
    \textbf{Disjointwith}: does-not-cause, prevent.
\end{itemize}

\subsection{Rules}
Throughout this section, we use four placeholders  -- \(A\), \(B\), \(C\), and \(D\) -- to represent events (or entities) that can be related by \textit{cause}, \textit{enable}, \textit{intend}, \textit{prevent}, or \textit{no-relation}. We consider the events \(A\) and \(B\) and their relationship “\(A \rightarrow B\)” in the claim and \(C\) and \(D\) with the relationship “\(C \rightarrow D\)” in the evidence. In the following subsections, we outline four key scenarios, all illustrated with examples.

\subsubsection{Logical alignment}
The \emph{logical alignment} scenario is verified if the claim and the evidence include the same (or similar, or transitively-linked) events, which are also connected by the same relation.
The evidence \textit{supports} the claim through logical alignment if the relation in claim and in evidence is the same and at least one of the following cases is verified:

\begin{itemize}
    \item \(C\) is similar to \(A\) and \(D\) is similar to \(B\); 
    \item a possible relation exists between \(A\) and \(C\) and/or between \(B\) and \(D\) which offer partial support by transitivity.
\end{itemize}
In other words, while similarity between events provides a clear pathway to alignment, a direct causal connection can also strengthen the claim in cases where event similarity is not established.

\textbf{Claim:} \textit{Sumo wrestler Toyozakura Toshiaki committed match-fixing, ending his career in 2011.}

\textbf{Evidence:} \textit{He was forced to retire in April 2011 after an investigation by the Japan Sumo Association found him guilty of match-fixing.}

\[
\begin{array}{rlrl}
A: & \textit{Committed match fixing} & 
C: & \textit{Investigation} \\
B: & \textit{Ending} & 
D: & \textit{Retire}
\end{array}
\]
\begin{enumerate}
    \item $A \xrightarrow{\text{causes}} B$ \hspace{0.5cm} (\textit{Direct cause from the claim})
    \item $C \xrightarrow{\text{causes}} D$ \hspace{0.5cm} (\textit{Direct cause from the evidence})
    \item $A \xrightarrow{\text{causes}} C$ \hspace{0.5cm} (\textit{Direct cause linking claim and evidence})
    \item $B = D$ \hspace{1.2cm} (\textit{Equivalence from the evidence})
\end{enumerate}

\textbf{Deduction:}
\begin{align*}
1. &\quad A \xrightarrow{\text{causes}} C \text{ and } C \xrightarrow{\text{causes}} D \implies A \xrightarrow{\text{causes}} D \quad \text{(by transitivity)} \\
2. &\quad B = D \implies A \xrightarrow{\text{causes}} B
\end{align*}

\begin{figure}[tbp]
 \centering
 \includegraphics[width=0.6\linewidth]{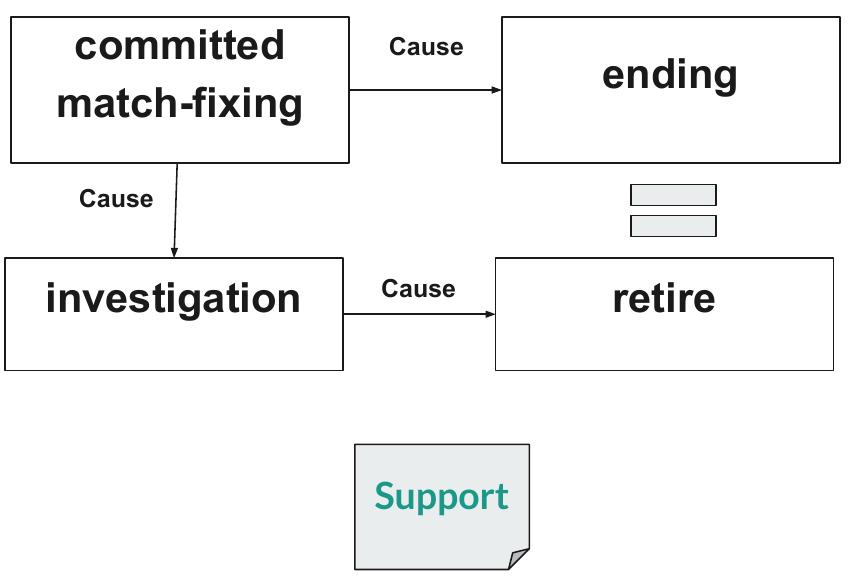}
 \caption{An example of a logical alignment.}
 \Description{An example of a logical alignment.}
 \label{fig:exp2}
\end{figure}

\textbf{Conclusion:}
$A \xrightarrow{\text{causes}} B$ is confirmed through transitivity (via evidence).
The evidence and the claim are logically consistent, demonstrating proper alignment. Figure \ref{fig:exp2} illustrates this scenario.

\subsubsection{Logical Misalignment}
In the \emph{logical misalignment} scenario, the relation in the evidence and the one in the claim can be opposite. If we find a similarity matching (\(C\) is similar to \(A\) and \(D\) is similar to \(B\)), we can conclude a direct contradiction to the claim: the same event cannot both cause (or enable/intend) and prevent the same outcome, making the evidence more likely to \textit{refute} the claim. 

\textbf{Claim}: \textit{Exercising daily causes muscle fatigue over time}.

\textbf{Evidence:} \textit{Research shows that daily low-intensity exercise  activates recovery mechanisms in the body, preventing the onset of chronic muscle fatigue and improving overall stamina instead.}

\[
\begin{array}{rlrl}
A: & \textit{Exercising daily} & 
C: & \textit{Activates recovery mechanisms} \\
B: & \textit{Muscle fatigue} & 
D: & \textit{Prevents muscle fatigue}
\end{array}
\]

\begin{enumerate}
    \item $A \xrightarrow{\text{causes}} B$ \hspace{0.5cm} (\textit{from the claim})
    \item $A \xrightarrow{\text{causes}} C$ \hspace{0.5cm} (\textit{from the evidence})
    \item $C \xrightarrow{\text{prevents}} D$ \hspace{0.3cm} (\textit{from the evidence})
    \item  $B = D$ \hspace{0.5cm} 
\end{enumerate}

\textbf{Deduction:}
Using the above relationships:
\begin{align*}
1. &\quad A \xrightarrow{\text{causes}} C \\
2. &\quad C \xrightarrow{\text{prevents}} D \implies C \xrightarrow{\text{causes}} \neg D \\
   &\quad \implies A \xrightarrow{\text{causes}} \neg D \implies A \xrightarrow{\text{prevents}} D \\
3. &\quad A \xrightarrow{\text{causes}} B \quad \text{and simultaneously} \quad A \xrightarrow{\text{prevents}} B \quad \text{(Contradiction)}.
\end{align*}

\textbf{Contradiction:}
A single event ($A$: \textit{Exercising daily}) cannot simultaneously \textit{cause} and \textit{prevent} the same outcome ($B$: \textit{Muscle fatigue}).

\begin{figure}[htbp]
 \centering
 \includegraphics[width=0.6\linewidth]{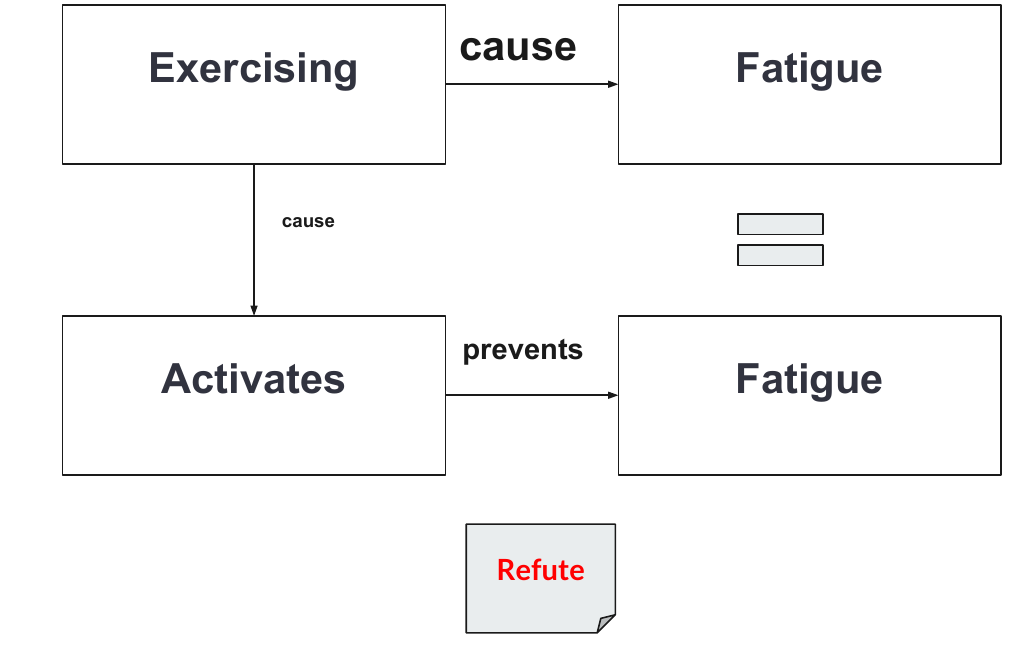}
 \caption{An example of a logical misalignment.}
 \Description{An example of a logical misalignment.}
 \label{fig:exp1}
\end{figure}

\textbf{Conclusion:}
The claim and evidence are logically inconsistent, as the evidence contradicts the claim's assertion (\figref{fig:exp1}).

\subsubsection{Causal loops}
We check for a \emph{closed causal loop} among four events \(A\), \(B\), \(C\), and \(D\) by looking at the relationships (cause, enable, intend, or prevent) between each pair. We first take a claim (\(A \rightarrow B\)) and an evidence (\(C \rightarrow D\)) and infers how \(A\) might relate to \(C\) and how \(D\) might relate to \(B\). If all four relationships form a consistent cycle (such as a chain of causes, enables, or intends), we have a closed causal loop, which implies a high probability that the evidence is \textbf{supporting} the claim. Since “prevent” is by definition considered the cause of not happening of another event, two consecutive “prevent” relations effectively become a “cause” because of the effect of a double negation. 

\textbf{Claim:} \textit{Poor infrastructure causes economic decline.}

\textbf{Evidence:} \textit{transportation inefficiencies leads to supply chain disruptions and reduced economic activity.}

\[
\begin{array}{rlrl}
A: & \textit{Poor infrastructure} & 
C: & \textit{Transportation inefficiencies} \\
B: & \textit{Economic decline} & 
D: & \textit{Supply chain disruptions}
\end{array}
\]

\begin{enumerate}
    \item $A \xrightarrow{\text{causes}} B$ \hspace{0.5cm} (\textit{from the claim})
    \item $A \xrightarrow{\text{causes}} C$ \hspace{0.5cm} (\textit{from the evidence})
    \item $C \xrightarrow{\text{causes}} D$ \hspace{0.5cm} (\textit{from the evidence})
    \item $D \xrightarrow{\text{causes}} B$ \hspace{0.5cm} (\textit{from the evidence})
\end{enumerate}

\textbf{Deduction:}
\begin{align*}
1. &\quad A \xrightarrow{\text{causes}} C \quad \text{and} \quad C \xrightarrow{\text{causes}} D \implies A \xrightarrow{\text{causes}} D  \\
2. &\quad A \xrightarrow{\text{causes}} D \quad \text{and} \quad D \xrightarrow{\text{causes}} B \implies A \xrightarrow{\text{causes}} B .
\end{align*}

\textbf{Conclusion:}
The claim $A \xrightarrow{\text{causes}} B$ is supported by the evidence through a causal loop. 

\subsubsection{Cherry-picking}
The practice commonly addressed as \emph{cherry-picking} consists of internal inconsistencies or selective usage of evidence. We group all evidence entries under the same claim, and then compares each pair of evidence elements. Each piece of evidence is represented as a \(\langle \text{sub}, \text{rel}, \text{obj} \rangle\) triple, where “\(\text{sub}\)” and “\(\text{obj}\)” are events or entities, and “\(\text{rel}\)” is the relationship between them. The code measures how similar these events/entities are (e.g. \(\text{sub}_1\) vs. \(\text{sub}_2\), \(\text{obj}_1\) vs. \(\text{obj}_2\)). 

A claim is flagged for cherry-picking if certain patterns in the evidence emerge. For instance, it checks whether two pieces of evidence use the \textbf{same relationship} (\(\text{rel}_1 = \text{rel}_2\)) but involve subjects or objects that are  dissimilar or opposites. If any of these mismatches is found, we deem the set of evidence potentially cherry-picked, because the evidence is either inconsistently presented or selectively used to reinforce the same relation in conflicting ways. 

\textbf{Evidence 1:} \textit{Frequent testing of the entire population would help identify so-called hidden carriers, individuals infected with SARS-CoV-2 [...] Identifying these silent spreaders could help public health workers be more effective at contract tracing by identifying others who have been exposed and may require quarantine.}

\textbf{Evidence 2:} \textit{Testing the entire population would undoubtedly identify a large number of such individuals, unnecessarily sidelining them from work and society.}

\begin{figure}[tbp]
 \centering
 \includegraphics[width=0.6\linewidth]{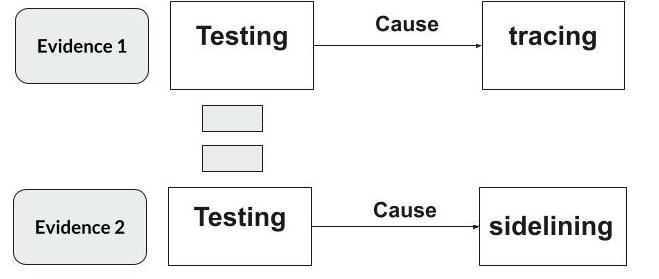}
 \caption{An example of a Cherry-Picking Scenarios.}
 \Description{An example of a Cherry-Picking Scenarios.}
 \label{fig:exp3}
\end{figure}

While the event \textit{Testing} from the first evidence is equivalent to \textit{Testing} from the second evidence, the second components of the relation differ significantly. Not only are they dissimilar, but they also have opposite polarities, the first being positive (tracing), while the second being negative (sidelining). This discrepancy may raise concerns about a potential cherry-picking scenario (\figref{fig:exp3}).

\section{Methodology}
\label{sec:reasoner}
\begin{figure}[htbp]
    \centering
    \includegraphics[width=1\linewidth]{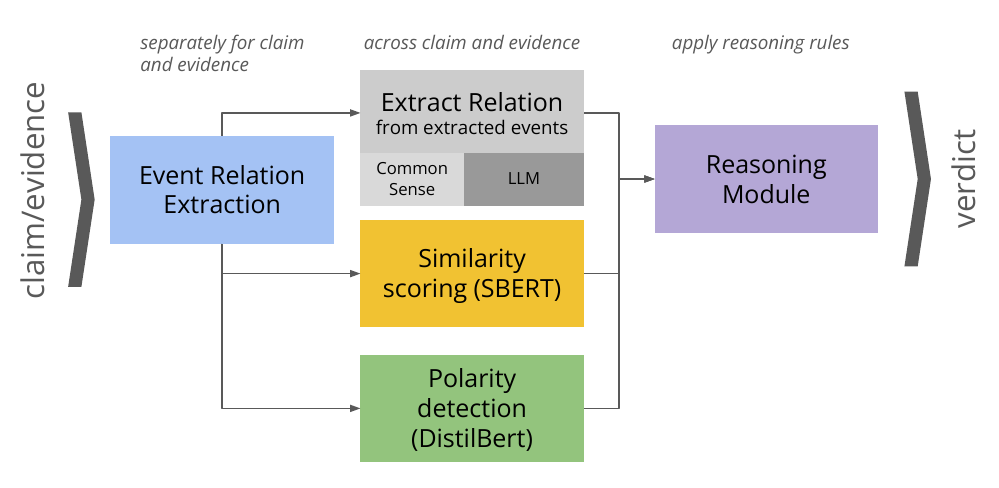}
    \caption{Overview of our proposed pipeline}
    \Description{Overview of our proposed pipeline}
    \label{fig:schema}
\end{figure}

Our pipeline (\figref{fig:schema}) begins with event relation extraction conducted separately within the claim and evidence (\secref{sec:extr-sep}). We describe the approaches used to extract fine-grained causal relationships across claims and evidence (\secref{sec:extr-acr}) and to distinguish between similar, dissimilar, and opposite events (\secref{sec:similarity}). These modules are then combined in our reasoning approach (\secref{sec:reasoning}).

\subsection{Causality Extraction within Claim/Evidence}
\label{sec:extr-sep}
Causality extraction is initially performed within the same context (either the claim or the evidence) using an automatic filtering process followed by manual validation. The system leverages the sequence-to-sequence model REBEL~\cite{huguet-cabot-navigli-2021-rebel-relation}, an auto-regressive architecture comprising an encoder and a decoder layer. We trained REBEL on a previously available annotated news dataset~\cite{GPTaugment} that includes 2,696 training, 265 development, and 461 testing sentences annotated with events and the four studied fine-grained causal relations. It processes raw text to produce the corresponding triplet \textit{(Subject, Relation, Object)}. This system obtained a combined F1-score of 0.82 (0.89 for identifying the subject, 0.75 for the object).

\subsection{Causality Extraction across Claim/Evidence}
\label{sec:extr-acr}
\begin{figure}[htbp]
 \centering
 \includegraphics[width=0.7\linewidth]{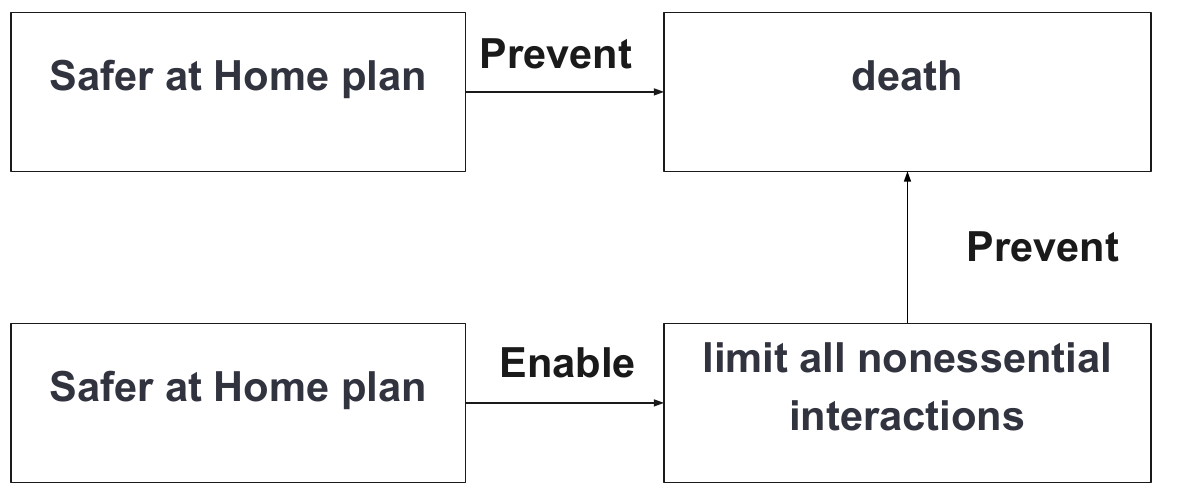}
 \caption{An example of refined causality extraction between events across the Claim and the Evidence}
 \Description{An example of refined causality extraction between events across the Claim and the Evidence}
 \label{fig:exp7}
\end{figure}

In \figref{fig:exp7}, we aim to identify the prevention relationship between the event \textit{``limit all non essential interactions''} in the evidence and the event \textit{``death''} in the claim. We have experimented with two different strategies: one relying on common sense knowledge bases and one relying on Large Language Models (LLM).

\subsubsection{Common Sense-based Causality Extraction}
The Atlas of Machine Commonsense (ATOMIC)~\cite{Sap2019ATOMICAA} is a large knowledge graph that contains over 877k inferential tuples representing common-sense situations, related by known relations. We identified an overlap between its relations and causality relations. Specifically, the relations \texttt{xIntent}/\texttt{oWant}, and \texttt{oEffect} are clearly expressing intention and direct cause. We isolated the ATOMIC tuples involving these properties, creating a reference dataset.

To address the absence of \textit{enabling} and \textit{prevention}, we have expanded the dataset by generating new examples using a LLM. Specifically, we used Zephyr, a fine-tuned version of Mistral 7B. As manual assessment, we randomly selected and judge 100 examples from each relation type (\textit{prevention}, \textit{enable}),  revealing that a significant portion of the generated examples (around 97\% for prevent and 76\% for enable) were accurate enough to train our model.

To handle the cases where there is no relation between the two events, we use negative sampling. We select 50\% of the whole dataset while stratifying the relation type, and we organize the dataset as a set of pairs of rows, so that a pair is two rows or two triples. Then, we swap randomly either the subject of the pair or the object. \tabref{tab:causality_extraction_atomic} shows the final support of the common sense dataset, and the positive results obtained on the test set.

\begin{table}[ht]
\centering
\begin{tabular}{|l|c|c|c|c|}
\hline
\textbf{Class}       & \textbf{Support} & \textbf{Precision} & \textbf{Recall} & \textbf{F1-Score} \\ \hline
\texttt{cause}   & 82,242    & 0.8248             & 0.8424          & 0.8335            \\ \hline
\texttt{intend}    & 146,588  & 0.8523             & 0.8924          & 0.8719            \\ \hline
\texttt{prevent}   & 53,454  & 0.9849             & 0.9929          & 0.9889            \\ \hline
\texttt{enable}   & 65,485   & 0.9755             & 0.9776          & 0.9765            \\ \hline
\texttt{no\_relation} & 173,886 & 0.8669            & 0.8208          & 0.8432            \\ \hline
\end{tabular}
\caption{Results of causality extraction between Claim Events and Evidence Events using ATOMIC augmented with LLMs}
\label{tab:causality_extraction_atomic}
\end{table}

\subsubsection{LLM-based Causality Extraction}
We experimented with the \textit{Phi-3-Medium-4K-Instruct} model, which demonstrates good performance in common-sense reasoning, while at the same time requiring fewer parameters and less computational effort~\cite{phi}. The following prompt was employed in a few-shot setting:

\begin{tcolorbox}[colframe=black, colback=white, coltitle=white, title=Event Relation Extraction Prompt]
\textbf{\texttt{User:}} Knowing that I want to extract refined causal relations between two given events, and it only can be \textit{cause}, \textit{intend}, \textit{prevent}, \textit{enable}, or \textit{no relation}, you have to answer only with the relation name, no explanation. What will be the relation between earthquake and death?

\textbf{\texttt{Assistant:}} cause.

\textbf{\texttt{User:}} What about relation between \{event1\} and \{event2\}?

\end{tcolorbox}

To evaluate the performance of the LLM in extracting fine-grained causality between events across claims and evidence, we extracted  and manually assessed 40 samples from both claims and evidences. Out of the 40 samples, 33 were correctly processed (82.5\%).

\subsection{Similarity, Dissimilarity, and Opposites}
\label{sec:similarity}
To determine if two events are the same, we rely on sentence similarity and dissimilarity, computed using SentenceBERT~\cite{reimers-gurevych-2019-sentence}.

We evaluate two configurations: (1) \textit{Events only} and (2) \textit{Events with context}, that is the concatenation of event spans and their original claim/evidence sentence. For cases where the events are an exact match (same surface form), we rely on the \textit{Events only} configuration, and otherwise apply \textit{Events with context}. To illustrate this approach, we use of the following examples.

\begin{enumerate}
    \item 
    \textbf{Claim:} \textit{Dr. Qadir went on hunger strike when in prison [...]; then he was released from custody on January 25, 2006, as a result of \textbf{efforts by special envoy of the Austrian foreign ministry}, Gudrun Harrer, a journalist}.\\
    \textbf{Evidence:} \textit{He was released from custody on January 25, 2006, as a result of \textbf{efforts} by special envoy of the Austrian foreign ministry, Gudrun Harrer.}
    \item 
    \textbf{Claim:} \textit{Sumo wrestler Toyozakura Toshiaki committed match-fixing, \textbf{ending} his career in 2011.}\\
    \textbf{Evidence:} \textit{He was forced to \textbf{retire} in April 2011 after an investigation by the Japan Sumo Association found him guilty of match-fixing.}

    \item
    \textbf{Claim:} \textit{The \textbf{drought} caused severe crop failures.} \\
    \textbf{Evidence:} \textit{Because of the prolonged dry \textbf{conditions}, agricultural yields were dramatically lower than usual.}
\end{enumerate}

In these examples, we have an alignment between events in the claim and events in the evidence. For each case, the \textit{Events with context} configuration produced the higher similarity score, as shown in \tabref{tab:similarity}. Based on empirical results on the entire dataset, we set the threshold for event similarity to \textbf{0.54}, as events with similarity above this value are considered similar.

\begin{table}[b]
\centering
\begin{tabular}{|c|c|c|c|}
\hline
\textbf{Input} & \textbf{Ex1} & \textbf{Ex2} & \textbf{Ex3} \\ \hline
Events only & 0.3235 & 0.2448 & 0.2589 \\ \hline
Events + context & \textbf{0.7632} & \textbf{0.6709} & \textbf{0.6533} \\ \hline
\end{tabular}
\caption{Cosine similarity for the three examples described above.}
\label{tab:similarity}
\end{table}

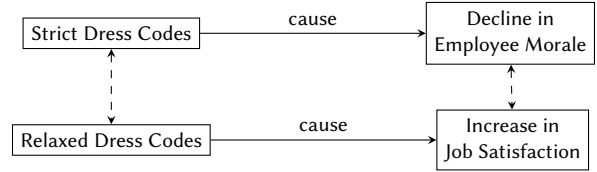
\begin{figure}[tbp]
\centering
\begin{tikzpicture}[%
    node distance=3cm,
    >=stealth,
    font=\small \sffamily
]

\node[draw, rectangle, align=center] (strict) {Strict Dress Codes};
\node[draw, rectangle, align=center, right=of strict] (decline) {Decline in\\Employee Morale};
\draw[->] (strict) -- node[above]{cause} (decline);


\node[draw, rectangle, align=center, below=1cm of strict] (relaxed) {Relaxed Dress Codes};
\node[draw, rectangle, align=center, right=of relaxed] (increase) {Increase in\\Job Satisfaction};
\draw[->] (relaxed) -- node[above]{cause} (increase);

\draw[dashed, <->] (strict) -- (relaxed);
\draw[dashed, <->] (decline) -- (increase);

\end{tikzpicture}
\caption{Two pairs of “opposite” triples.}
\Description{Two pairs of “opposite” triples illustrating cause relationships.}
\label{fig:opposites}
\end{figure}

Sometimes events represent concepts that are simply \emph{dissimilar} (indicated by a similarity score falling below a certain threshold), while in other cases, they represent exact \emph{opposites}. According to \cite{CRUTCH20122636}, antonyms can be as similar or even more similar than synonyms, aside from their polarity (Pol). This insight suggests that we can detect opposites by identifying pairs of events with high similarity but contrasting polarities.

To investigate this, we sampled five pairs of claims and corresponding evidence in which the statements contradict each other, and the events involved are opposites. We computed similarity and polarity under three configurations:
\begin{itemize}
    \item Evaluating events in isolation,  
    \item Evaluating whole claim text vs whole evidence text, and  
    \item Evaluating the full triple text \textit{(sub,rel,obj)}.
\end{itemize}

We discuss the following example, represented in \figref{fig:opposites}:\\
\textbf{Claim:} \textit{A study released on November 15, 2023, found that companies with strict dress codes experience a decline in employee morale.}\\
\textbf{Evidence:} \textit{The Workplace Institute surveyed 150 firms with relaxed dress codes in October 2023 and found that employees reported a 25\% higher job satisfaction rate than those at 100 companies enforcing formal attire. 
} \\

We determined which configuration best captured the correct polarity alongside high similarity. The configuration that produced the correct opposing polarities with strong similarity scores was ultimately chosen to identify and confirm opposite relationships. 

Similarity is computed as the cosine similarity between the embeddings of two input events obtained from SentenceBERT.
For polarity detection, we employed the DistilBERT base uncased model fine-tuned for sentiment analysis\footnote{\url{https://huggingface.co/distilbert/distilbert-base-uncased-finetuned-sst-2-english}}.
The model takes the input text and outputs a polarity classification: \(N\) (Negative) or \(P\) (Positive).

Evaluating the complete triple yields the highest percentage of correct polarities, and the second highest similarity scores making it the most effective among the considered configurations (\tabref{tab:comparison}). 

Based on the empirically determined similarity threshold $\theta$ and the opposites check, which suggests that opposites are semantically similar but exhibit different polarities, we define the following rules. Given two text inputs \( T_1 \) and \( T_2 \):

\begin{itemize}
    \item \textbf{Is-Similar}(\( T_1, T_2 \)) \(\implies\) \\ \(\text{Similarity}(T_1, T_2) > \theta\) \textbf{and} \(\text{Pol}(T_1) = \text{Pol}(T_2)\) (\(PP\) or \(NN\)).
    \item \textbf{Is-Dissimilar}(\( T_1, T_2 \)) \(\implies\) \(\text{Similarity}(T_1, T_2) < \theta\).
    \item \textbf{Opposites}(\( T_1, T_2 \)) \(\implies\) \\ \(\text{Similarity}(T_1, T_2) > \theta\) \textbf{and} \(\text{Pol}(T_1) \neq \text{Pol}(T_2)\) (\(PN\) or \(NP\) )
\end{itemize}

\begin{table}[ht]
    \centering
    \begin{tabular}{lcc}
        \toprule
        \textbf{} & \textbf{Correct} & \textbf{Average} \\
        \textbf{Comparison} & \textbf{Polarity} & \textbf{Similarity} \\
        \midrule
        Claim vs Evidence & 20\% & \textbf{0.76} \\
        Claim Event vs Evidence Event & 50\% & 0.50 \\
        Triple Claim vs Triple Evidence & \textbf{80\%} & 0.61 \\
        \bottomrule
    \end{tabular}%
    \caption{Comparison of correct polarity and average similarity across different levels of analysis.}
    \label{tab:comparison}
\end{table}

\subsection{Reasoning Approach}
\label{sec:reasoning}
\begin{figure}[htbp]
\centering
    \centering
    \includegraphics[width=\linewidth]{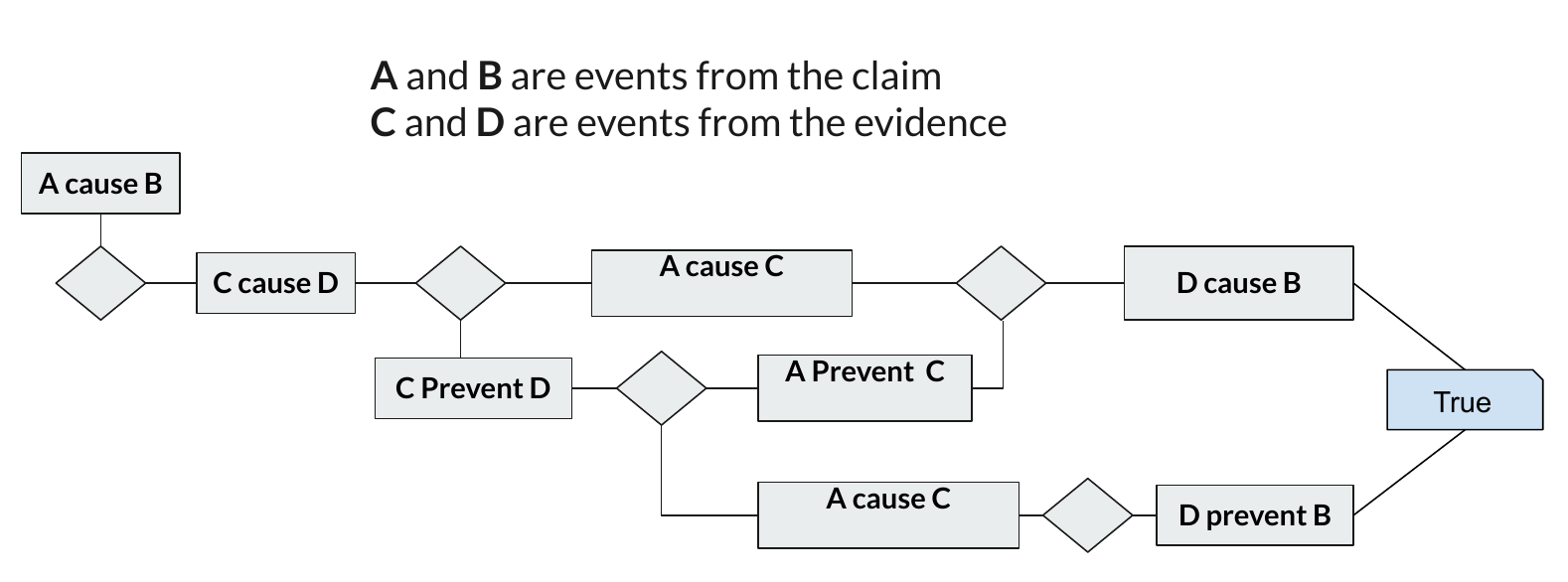}
    \caption{Implementation structure of the Causal Loop check.}
    \Description{Implementation structure of the Causal Loop check.}
    \label{fig:exp4}
\end{figure}

\begin{figure}[htbp]
    \centering
    \includegraphics[width=\linewidth]{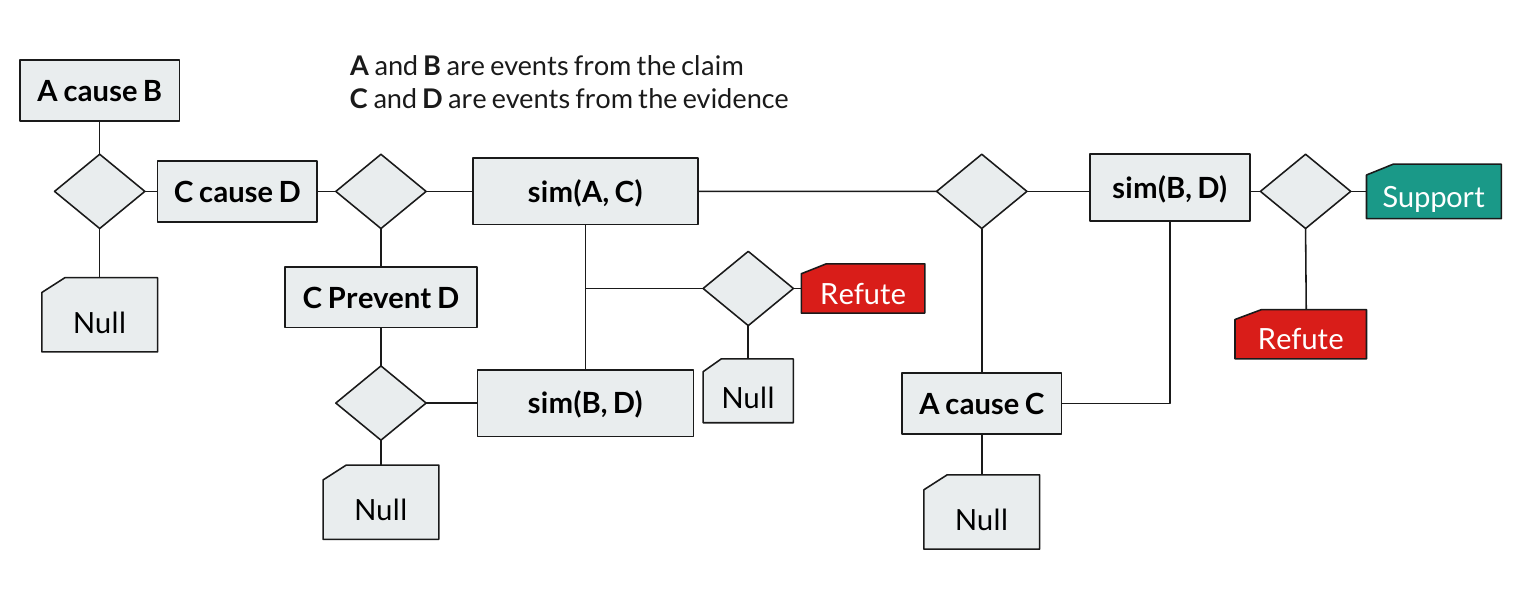}
    \Description{Implementation structure of a Similarity and Relationship check. \textit{Is-similar} is shortened to \textit{sim} for readability} 
    \caption{Implementation structure of a Similarity and Relationship check. \textit{Is-similar} is shortened to \textit{sim} for readability} 
    \label{fig:exp5}
\end{figure}

\begin{figure}[htbp]
\centering
\includegraphics[width=\linewidth]{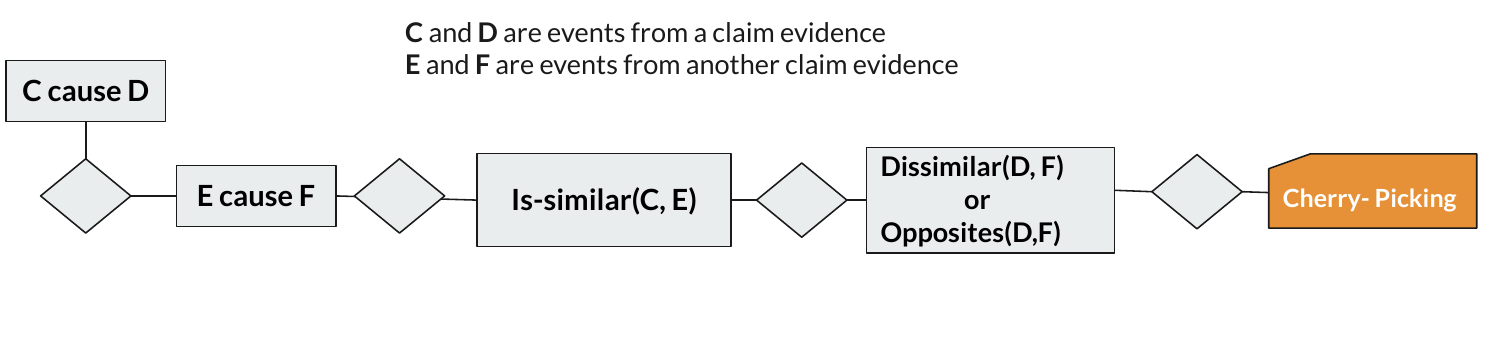}
\caption{Implementation structure of a cherry picking scenario check.}
\Description{Implementation structure of a Cherry picking check.}
\label{fig:exp6}
\end{figure}

The reasoner analyzes the claim and evidence by checking the relationships between their events. It performs the following tasks:
\begin{enumerate}
    \item \textbf{Causal Loop check} (\figref{fig:exp4}) verifies if the events form a closed causal cycle, indicating support for the claim.
    \item \textbf{Similarity and Relationship check} (\figref{fig:exp5}) compares the relationships and similarities between events to determine alignment or contradiction. 
    \item \textbf{Cherry-picking check} (\figref{fig:exp6}) identifies inconsistencies or selective usage of evidence that may bias the verdict.
\end{enumerate}


The assigned labels are categorized as in \textsc{AVeriTeC} (see \secref{sec:evdat}):
\begin{itemize}
    \item \textbf{Supported:} When the evidence fully supports the claim.
    \item \textbf{Refuted:} When the evidence contradicts the claim.
    \item \textbf{Conflicting Evidence/Cherrypicking:} When the evidence neither fully supports nor completely refutes the claim, but exhibits conflicting information.
\end{itemize}

\section{Evaluation}
\label{sec:eval}
This section outlines the evaluation datasets and the strategies used to assess our reasoning framework. We explain the filtering criteria applied to each dataset and the evaluation setups adopted for performance analysis.

\subsection{Evaluation Datasets}
\label{sec:evdat}
To assess the effectiveness of our reasoning framework, we conducted evaluations on two widely used fact-checking datasets: \textsc{AVeriTeC}~\cite{schlichtkrull2023averitec} and \textsc{FEVEROUS}~\cite{Aly21Feverous}.
Additionally, we constructed a manually curated subset of relevant use cases.

\subsubsection{AVeriTeC}
It consists of 4,568 real-world claims, each paired with question--answer evidence and textual justifications used to determine verdicts, and annotated with one of four verdict labels:
\begin{itemize}
    \item \textbf{Supported:} The evidence fully supports the claim.
    \item \textbf{Refuted:} The evidence directly contradicts the claim.
    \item \textbf{Conflicting Evidence (Cherry-picking):} The evidence presents conflicting information neither fully supporting nor fully refuting the claim.
    \item \textbf{Not Enough Evidence}: The evidence is insufficient to make a conclusive judgment about the claim veracity. 
\end{itemize}

In our evaluation, we use the training subset of \textsc{AVeriTeC}, but we retained only claims linked to informative textual answers, so excluding boolean and unanswerable cases. We also exclude the \textit{Not Enough Evidence} portion of the dataset, since our reasoning system is not designed to produce this type of verdict. This leaves 2998 claims.

\subsubsection{FEVEROUS dataset}
It contains 87,026 claims annotated with evidence sourced from Wikipedia. Each claim is labelled as \textit{supports}, \textit{refutes}, or \textit{not enough info}. Evidence may include textual sentences or table cells, along with annotator metadata -- e.g., query actions, page clicks, evidence types.

We randomly selected a subset of 4,392 claims for our experiments, to have a dataset similar to the \textsc{AVeriTeC} in volume. We retained only claims supported by fully textual evidence and excluded those referencing table cells, yielding a filtered subset suitable for text-only based reasoning. We also excluded the \textit{Not enough info} part since it is not handled by the reasoner. We then filtered out claims without causal relations, retaining a total of 1,183 claims for evaluation (705 \textit{supports} and 478 \textit{refutes}).


\subsubsection{Reasoner Specific Subset (RSS)}
Our reasoning framework is based on a set of rule-based mechanisms operating over event relations, similarity, and polarity. While this approach allows for explicit and interpretable inference, it does not guarantee coverage of all examples within the datasets, as not all claim-evidence pairs contain use cases compatible with the system's reasoning rules.

As additional evaluation, we constructed a controlled subset consisting of claim-evidence pairs that contain verified use cases:
while we do not assume that the reasoner will always respond correctly (the final outcome also depends on other components, e.g. similarity scoring and polarity detection), we ensure that a valid use case is present and the mechanism should in principle be activated.

We randomly sampled 765 claims from the \textsc{AVeriTeC} dataset. For each claim-evidence pair, we checked whether it contained a valid use case for reasoning, e.g., causal loops or contradictions. We only include in this dataset the 86 pairs (across 60 unique claims) presenting a valid use case.

\tabref{tab:merged_filtering_steps} shows the stats for each dataset after filtering.

\begin{table}[h!]
\centering
\begin{tabularx}{\linewidth}{|X|c|c|}
\hline
\textbf{Filtering Step} & \textbf{\textsc{AVeriTeC}} & \textbf{\textsc{FEVEROUS}} \\ \hline
Total unique claims & 2998 & 1736 \\
Total answers / total evidences & 8479 & 3836 \\ \hline
\multicolumn{3}{|c|}{\textbf{Answer Type Distribution (AVeriTeC)}} \\ \hline
Extractive & 4571 & -- \\
Abstractive & 2225 & -- \\
Boolean & 1297 & -- \\
Unanswerable & 386 & -- \\ \hline
Claims \textit{\scriptsize excluding Boolean and Unanswerable} & 2783 & 1736 \\
Claims with no relation & 850 & 44 \\
Claims \textit{\scriptsize excluding "not enough evidence"} & 1759 & 1183 \\ \hline
\multicolumn{3}{|c|}{\textbf{Label Distribution}} \\ \hline
Refuted / REFUTES & 1066 & 478 \\
Supported / SUPPORTS & 581 & 705 \\
{\small Conflicting Evidence / Cherrypicking} & 112 & -- \\ \hline
\end{tabularx}
\caption{Filtering steps and label distributions for the datasets before running the reasoning pipeline}
\label{tab:merged_filtering_steps}
\end{table}

\subsection{Evaluation Strategy}
In our evaluation, we define two distinct configurations for computing performance metrics:
\begin{itemize}
    \item \textbf{Configuration 1 – Tolerant:}  
    It adopts a lenient evaluation strategy, focusing only on the system's performance when it chooses to respond.
    \begin{itemize}
        \item \textbf{Recall} is defined as the proportion of cases where the reasoner successfully produces a verdict (i.e., does not abstain), relative to the total number of evaluated cases.
        \item \textbf{Precision} measures the proportion of correct verdicts among those that were actually produced (i.e., abstentions are excluded).
        \item \textbf{Abstentions} (\textit{None} outputs) are excluded from metric computation. This is because they may arise from either genuinely irrelevant inputs or from missed reasoning opportunities due to limitations in similarity matching or claim--evidence alignment. 
    \end{itemize}

    \item \textbf{Configuration 2 – Strict:}  
    This configuration uses a more rigorous evaluation policy. It is primarily applied to the manually verified RSS, where each item has been checked to ensure that it is reasoning-relevant.    
    \begin{itemize}
        \item The system is expected to always produce a verdict. Every abstention (\textit{None}) is treated as a \textbf{false negative (FN)}, thus reducing recall.
        \item \textbf{Recall} is computed as the number of correct verdicts over the total number of evaluated cases, including abstentions.
    \end{itemize}
\end{itemize}
In \textsc{AVeriTeC} and \textsc{FEVEROUS}, all examples—including abstentions—are evaluated, with abstentions counted as false negatives. This strict setup tests the model’s robustness in ambiguous cases.

\subsection{Results and Discussion}
\label{sec:result}

\begin{table}[b]
\centering
\renewcommand{\arraystretch}{1.2}
\setlength{\tabcolsep}{6pt}

\begin{tabularx}{\linewidth}{|X|X|c|c|c|}
\hline
\textbf{Test Set} & \textbf{Knowledge Source} & \textbf{P} & \textbf{R} & \textbf{F1-Score} \\ \hline
\multirow{2}{*}{\textbf{RSS}} & LLMs & 0.55 & 0.45 & 0.50 \\ \cline{2-5}
& Common Sense & 0.51 & 0.45 & 0.48 \\ \hline
\multirow{2}{*}{\textbf{\textsc{AVeriTeC} (S)}} & LLMs & 0.48 & 0.19 & 0.27 \\ \cline{2-5}
& Common Sense & 0.54 & 0.2 & 0.29 \\ \hline
\multirow{2}{*}{\textbf{\textsc{AVeriTeC} (T)}} & LLMs & 0.47 & 0.35 & 0.4 \\ \cline{2-5}
& Common Sense & 0.52 & 0.37 & 0.43 \\ \hline
\multirow{2}{*}{\textbf{\textsc{FEVEROUS} (S)}} & LLMs & 0.5 & 0.44 & 0.47 \\ \cline{2-5}
& Common Sense & 0.51 & 0.44 & 0.47 \\ \hline
\multirow{2}{*}{\textbf{\textsc{FEVEROUS} (T)}} & LLMs & 0.52 & 0.62 & 0.56 \\ \cline{2-5}
& Common Sense & 0.52 & 0.62 & 0.56 \\ \hline
\end{tabularx}
\caption{Precision, recall, and F1-score for each knowledge source across the different evaluation datasets. RSS refers to the Reasoner-Specific Subset, in which tolerant evaluation was unnecessary as all examples are guaranteed to trigger reasoning. (S) stands for the Strict evaluation while (T) refers to the Tolerant one.}
\label{tab:reasoner_performance}
\end{table}

\tabref{tab:reasoner_performance} reports the performance of our reasoning framework across different configurations, datasets, and knowledge sources for causality extraction across claims and evidences. On the reasoner specific subset (RSS), the system achieves an F1-score of 0.50 with LLMs and 0.48 with common-sense knowledge bases ERE. In 50\% of the cases the model either does not provide an answer or the answer is wrong.

Upon analyzing the failure cases in the \textsc{RSS} dataset, we observe that the Event Relation Extraction component exhibits notable inconsistencies. Some events are overly abbreviated, while others lack essential lexical content, making subsequent similarity and polarity computations highly unreliable.

Moreover, the model struggles with complex linguistic structures, such as double negation. For instance, given the claim \textit{(drinking water, intend, protect covid)} and the evidence 
\textit{(hydrated, does not cause, coronavirus infection)},
the model fails to resolve the logical equivalence due to nested negation and lexical variation.

Another recurrent source of error stems from the reasoner’s current inability to perform type-based or ontological reasoning.
For example, with the claim \textit{(5G, causes, infertility)} against the evidence \textit{(non-ionizing radiation, does not cause, infertility)}, the model is unable to infer that 5G, is a form of non-ionizing radiation, and shares the relevant properties. In the absence of explicit contextual or ontological information, the system fails to capture the semantic similarity required for correct inference.

The evaluation of \textsc{AVeriTeC} and \textsc{FEVEROUS} reveals notable differences in performance between strict and tolerant settings, offering distinct insights into how each dataset responds to these configurations. For \textsc{AVeriTeC}, the strict setting leads to a sharp decline in recall for both sources, a result that aligns with expectations since abstentions—classified as "None" are treated as false negatives. Conversely, the tolerant configuration paints a more favorable picture: when the system does provide a response, its predictions tend to be accurate, as reflected in the higher F1-scores.
In the case of \textsc{FEVEROUS}, performance remains more consistent across both reasoning approaches. Under the strict setting, both reasoners achieve an F1-score of 0.47, while the tolerant setting sees this figure rise to 0.56 for both, with recall reaching as high as 0.62. This pattern indicates that \textsc{FEVEROUS} likely contains more straightforward, fact-based claim-evidence pairs, easier to align and reason about, resulting in more stable performance across configurations.

\paragraph{Comparison with GPT-4o mini.} 
To better assess our work, we compared our rule-based reasoner against a powerful large language model (LLM), GPT-4o mini, for the RSS dataset. GPT-4o mini was provided with each claim together with the complete set of corresponding evidences as input. The model achieved a macro F1-score of 0.67 across the three labels, outperforming our reasoner, which reached a score of 0.50. This corresponds to an improvement of approximately +17\%. However, GPT-4o mini may have been previously exposed to similar samples from the RSS dataset during training, a contamination that could lead to an overestimation of its real performance. Despite its advanced reasoning abilities, the model still struggles to achieve perfect consistency, confirming the intrinsic complexity of causal relation reasoning in factual verification.

\section{Conclusion and Future Work}
\label{sec:conclusion}
This study presents a novel approach for incorporating causal reasoning into automated fact-checking. By leveraging semantically refined event relationships and a structured reasoning framework, our system addresses the prevalent limitation of causal interpretability in existing methods—moving beyond vague or shallow representations of causality.

While the proposed reasoner achieves an F1-score of approximately 50\%, its main contribution is not in outperforming existing models in terms of raw accuracy, but in providing structured and interpretable justifications for fact-checking verdicts while still being very competitive. Rather than functioning as a standalone predictor, the system has the potential to complement existing black-box veracity classifiers, including those offering limited explainability based on vague or underspecified causal links. By surfacing explicit causal links, polarity mismatches, and logical inconsistencies, it provides valuable explanatory signals to support or challenge automated decisions.

Our error analysis highlights several avenues for improvement as future work. First, inconsistencies in ERE outputs, such as incomplete or overly abstract event representations significantly hinder downstream similarity and polarity computations. Future work will focus on improving event extraction robustness, potentially by integrating event typing. The event representation should also aggregate the information about time and space.

The current system struggles with complex linguistic phenomena such as double negation and implicit entailment. Incorporating symbolic reasoning layers or transformer-based inference modules fine-tuned on logical patterns could help address this limitation.

Finally, the system lacks the capacity for ontological reasoning, which is crucial for handling type-based mismatches (e.g., recognizing that 5G is a subclass of non-ionizing radiation). Future directions include enriching the model with external ontologies or commonsense knowledge bases to support type inference and contextual disambiguation. Such enhancements would improve both the coverage and reliability of the causal reasoning pipeline in real-world fact-checking scenarios.

\begin{acks}
This work was supported by the European CHIST-ERA program within the ClimateSense project (Grant n°  ANR-24-CHR4-0002, EPSRC EP/Z003504/1) and by the French National Research Agency (ANR) within the kFLOW project (Grant n° ANR-21-CE23-0028).
\end{acks}

\printbibliography

\end{document}